# Core-set Selection Using Metrics-based Explanations (CSUME) for multiclass ECG

Sagnik Dakshit
*Computer Science*
*The University of Texas at Dallas*
Dallas, USA
sdakshit@utdallas.edu

Barbara Mukami Maweu
*Computer Science*
*The University of Texas at Dallas*
Dallas, USA
barbara.maweu@gmail.com

Sristi Dakshit
*Computer Science*
*Calcutta Institute of Engineering and Management*
Kolkata, India
sristidakshit@gmail.com

Balakrishnan Prabhakaran
*Computer Science*
*The University of Texas at Dallas*
Dallas, USA
bprabhakaran@utdallas.edu

*Abstract*— The adoption of deep learning-based healthcare decision support systems such as the detection of irregular cardiac rhythm is hindered by challenges such as lack of access to quality data and the high costs associated with the collection and annotation of data. The collection and processing of large volumes of healthcare data is a continuous process. The performance of data-hungry Deep Learning models (DL) is highly dependent on the quantity and quality of the data. While the need for data quantity has been established through research adequately, we show how a selection of good quality data improves deep learning model performance. In this work, we take Electrocardiogram (ECG) data as a case study and propose a model performance improvement methodology for algorithm developers, that selects the most informative data samples from incoming streams of multi-class ECG data. Our Core-Set selection methodology uses metrics-based explanations to select the most informative ECG data samples. This also provides an understanding (for algorithm developers) as to why a sample was selected as more informative over others for the improvement of deep learning model performance. Our experimental results show a 9.67% and 8.69% precision and recall improvement with a significant training data volume reduction of 50%. Additionally, our proposed methodology asserts the quality and annotation of ECG samples from incoming data streams. It allows automatic detection of individual data samples that do not contribute to model learning thus minimizing possible negative effects on model performance. We further discuss the potential generalizability of our approach by experimenting with a different dataset and deep learning architecture.

## I. Introduction

The superiority of Deep Learning (DL) networks in providing decisions with high accuracy have led to an increase in the use of healthcare decision support systems such as Arrhythmia detection. The limited availability of data due to the high cost of collection, expert annotation, and poor quality hinders the performance of machine learning algorithms in life-critical domains such as healthcare. The importance of memory-efficient, data-quantity driven learning has been highlighted by Plutowski et. al [13], by establishing the proportional relation between the number of training examples with the model complexity. Synthetic data [1, 10] has been used as a supplementary source for data augmentation, at the risk of additional bias, noise, and wrong annotation, which shows the importance of data quantity in deep learning training. In this work, we highlight the influence of data quality on deep learning model performance and provide a Core-Set Selection strategy to maximize performance using metrics-based explanations.

*Core-Set Selection:* Core-set selection is the process of selecting a subset of samples from a pool of available samples. This selection assumes a labeled dataset and the derived subset is expected to retain the same or comparable model error to that achieved when training with an entire set. We show Core-Set selection as $S' = P_{labeled} - S_0$ where $S'$ is the subset of samples retained after removal of subset $S_0$ from the set $P_{labeled}$ which implies $|S'|<|P_{labeled}|$ where $P_{labeled}$ is a pool of labeled data. Coleman et. al [2] defines it as finding a subset $|S'|<=|S|$ of data samples where $P_{selected} = \{\{(x_j, y_j) \in \#N_{labeled}, j \in S'\}\}$ such that any model trained on subset $P_{selected}$ achieves similar or comparable accuracy to a model trained on $S$ samples. In our representation, $P_{selected}$ is the set of selected samples and $\#N_{labeled}$ number of samples in $P_{labeled}$.

*Interpretable Explanations:* Our proposed Core-Set learning is conditioned on an oracle that is defined as a trained supervised learner whereby generated explanations from the oracle are used to select a training subset from new data, that when added to the pool of training data, maintains, or improves the performance of the model. The focus of such Explainable Artificial Intelligence (XAI) methods can be categorized as Visual, Textual, and Metric based. The medium of explanation is specific for a set of intended users' interpretability. While visual and textual forms of explanations are easier to comprehend, metric-based explanations can help in model improvement due to their quantifiable nature. We use quantified metric forms of interpretable explanations as features to select the most influential samples from a dataset in improving the deep learning model performance. This consequently bolsters the evaluation of models for developers with interpretable reasons for the selection or rejection of certain samples in a dataset.

### A. Proposed CSUME Framework

We propose a novel algorithm, Core-set Selection Using Metrics-based Explanations (CSUME), to efficiently train DL decision support models that improve model performance in arrhythmia detection. Our proposed CSUME framework asserts the quality and annotation of the new incoming non-asserted dataset. Our framework is a Core-Set selection method that uses an algorithm based on quantifiable interpretable model explanations to select the most informative model training samples that demonstrate improvement in model performance iteratively, as illustrated in Figure 1. The quality of the non-asserted data is validated through our framework by excluding those data samples that impact model performance negatively. Furthermore, our framework allows an iterative model improvement on the selected samples from an incoming stream of data. This iterative nature allows for optimized, generalized performance and an up-to-date model. Through our experimental results in Section 4, we observe significant performance improvement in a model trained on data selected using our proposed algorithm over a model trained on the entire non-asserted data.

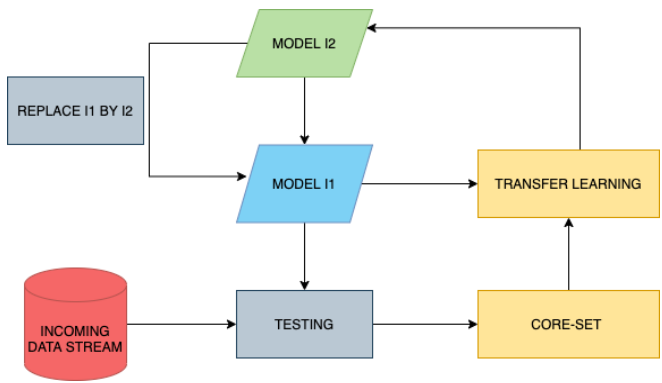

Fig. 1. (CSUME) Iterative Framework Flow; Model $I_i$ represents iteration number such as $I_1, I_2..I_n$

Core-set selection assumes access to a labeled dataset. It is the process of selecting a subset of the labeled dataset. As a use case, we demonstrate our proposed CSUME framework with electrical conduction abnormality condition, Arrhythmia. Arrhythmia, one of the most common critical heart conditions, bolsters the criticality of data quality. We maintain that the Core-Set samples achieve comparable or better model performance than a model trained with the entire dataset crucial need for maximizing performance for Arrhythmia decision support systems. Our proposed CSUME framework is intended for the ML expert for improvement of the system performance and not for the end user.

*D. Contributions*

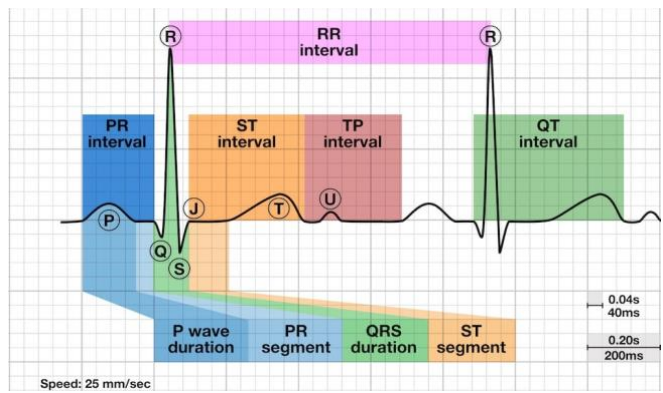

Fig. 2. ECG signal clinical peaks and segments [9]

We demonstrate the efficacy of our framework using Electrocardiograms (ECG) as the sample use case. ECG signals are bio-electrical patterns representing the heartbeat recorded as a 1D time-series waveform. The cardiac rhythm is a repetition of heart beats with distinctive features as shown in Figure 2. The segments and peaks of both intra-beat and inter-beat are used to diagnose medical conditions. We categorize ECG signals as: 1) Asserted *Data* for the acclaimed publicly available labeled ECG dataset used in research. The established nature of these ECG datasets makes them well trusted, and 2*) Non-Asserted Data* for newly generated or collected ECG datasets whose utility has not been established through research or by domain experts. The use of non-asserted data can potentially have a derogatory effect on model performance especially in life-critical systems where data assertion in terms of data quality and annotation is important.

Our contributions can be summarized as follows:

- Our proposed CSUME methodology uses metrics-based explanations for new, incoming, non-asserted data streams, as a tool for algorithm developers to improve DL models.
- Our proposed CSUME framework is iterative in nature and performs Core-Set selection on a continuous incoming data stream, keeping the model updated. Iteratively learning via selected Core-Set allows the decision support system to be state-of-the-art by maximizing model performance and minimizing model degradation due to outliers, noisy samples, or non-informative data by 2.05% in our experiments on a small dataset.
- We show a 9.67% model performance improvement with a reduction of 50% of the data.
- CSUME also performs Data Quality Assertion, the task of asserting the quality of the dataset and the associated annotations, for new, incoming, non-asserted data streams. This is done by identifying and removing data samples that have a derogatory effect on model performance owing to noise, skewness, or wrong annotation.
- We also perform an ablation study to understand the importance and ranking of each interpretable explanation metric.

II. RELATED WORKS

*A. Core-Set Selection Strategies*

The challenge of small datasets, high cost, irregular measurements, and expensive data annotation in the healthcare domain requires creative and effective methods of training classification learners. To address these challenges, researchers have explored methods that maximize outcomes of classification tasks when small quantities of model training data are available such as data mining [17], feature selection and pre-processing [8], Active Learning, and Core-Set selection [7, 5, 3]. An explanation based active learning method was proposed by [4]. Our proposed approach is in the domain of Core-Set selection where the dataset is labeled unlike Active Learning. Active Learning is primarily focused on the selection of samples to reduce annotation cost while Core-Set selection focuses on minimizing computational resources while maximizing model performance. The approach close to our work is [14] where the authors define an active learning problem as a Core-Set selection problem. They apply the min-max facility location problem to select an effective training subset. Similarly, our work strives to achieve a solution to better utilize small datasets using a Core-Set selection approach. Our work differs from [14] in that we propose applying quantified explanations derived from a CNN trained model to inform on the choice of the most informative and learnable subset. Furthermore, their work focuses on 2D image large-scale datasets such as CIFAR-10, CIFAR-100. We present a detailed comparison with [14] random and our method in Section 5.

Additionally, the approach by [2] also resonates close to our work. Their proposed use of proxy models does not work efficiently with structured time series data like ECG signals. This is evidenced in the inability of neural network architectures to capture the temporal and spatial features of the raw time-series signals, therefore, making it impossible to train proxy models which abstract layers and reduce convolution sizes. Without layer abstraction, the approach proposed by [2] is essentially a bootstrapping method. While their proposed approach sets a budget for the number of selected samples, our approach does not put a constraint on the number of samples but instead chooses the minimum number of samples required to maximize model performance. Furthermore, their approach focuses on minimizing the error and not on the precision and recall for individual classes which is crucial to understand the true capacity of a trained model used in healthcare diagnosis.

### B. Explainability in Time-Series Deep Learning

Explanations in literature are primarily classified as interpretable models and as post-hoc model explanations. Explainability for deep learning models can also be grouped based on the medium of explanation namely: 1) Visual, 2) Textual and 3) Quantifiable Metric. Motivated by the easier comprehension of humans, research on explainability has been primarily focused on Visual and Textual forms for transparency and accountability for wider social acceptance. Furthermore, Explainable Artificial Intelligence (XAI) has been more focused on image and tabular data modality with techniques such as LIME, GRADCAM [20,19] and leaving a gap in explanations for time-series signal data. Strum et. al [16] use Layer wise Relevance Propagation (LRP) to analyze time-series 1D electroencephalogram (EEG) data as an interpretable model. This method is however limited to assigned relevance scores of data and their contribution to prediction without accounting for the knowledge and features of the signals learned by the model. Post-hoc time-series explanations use extracted shapelets [18, 6] which are suited for pattern discovery that represents the target class. Karlsson et. al [6] proposed a non-DL method to provide explanations for time-series. Time-series tweaking is the minimum number of changes needed to change the classification result in a random forest ensemble. Research in explainability for time series is still in its early stages and most of the existing work cannot be used as a quantifiable metric for evaluation of model performance.

### C. Metrics-based Explainability for Time-Series

Metric-based explanations quantify clinical features used in diagnosis as shown by [9]. While [9] takes a step towards making neural networks gray-box by quantifying the capacity of a model using explainable metrics. In this paper, we address the other predominant healthcare challenge of scarcity and limited access to sensitive patient medical data, and high cost of medical data annotation. CSUME Framework uses such explainable metrics to improve deep learning performance by core-set selection. These metric-based explanations are intended for the algorithm developers and not for end-users.

Traditional performance measures such as accuracy, precision, and recall do not adequately evaluate the capacity of a model. Metrics have been used to generate interpretable explanations for model capacity using features learned by a model [9]. For use of explanations in Core-Set selection, quantification of the explanations is of utmost importance. Existing explanations methods for black-box deep learning systems present explanations that are not quantifiable. The above reasons motivate the use of Dynamic Time Warping (DTW), Mean Squared Error (MSE), and SLACK[1] as quantifiable metrics for 1D time-series interpretable explanations as established by Maweu et. al. Their results show that these metrics adequately represent model capacity and learned features specifically for 1D signal data such as ECG with comprehensively quantifying structural, frequency and clinical features. In this Section, we provide a brief overview of these metrics introduced by Maweu et. al [9].

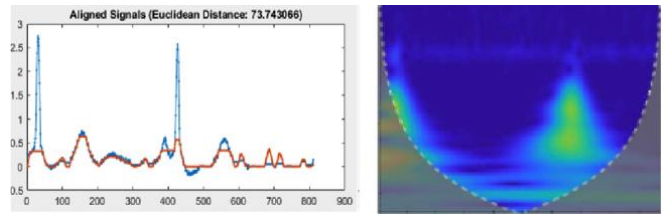

Fig. 3. Interpretable model explanation metrics used to evaluate learned features and capacity of model. X-axis represents time in ms and Y-axis is amplitude a) DTW over input ECG waveform and model feature map (b) CWT of ECG waveform.

### III. CSUME FRAMEWORK EXPERIMENTS AND RESULTS

**DTW:**
Dynamic Time Warping is a signal analysis algorithm that measures the similarity between any two given signals. DTW is shown to be a descriptive statistical explanation for ECG signals [9]. This helps quantify inherent features such as the shape of waveform. It is quantified as an interpretable metric between the representation of learned structural feature maps from any 1D CNN layer and the corresponding 1D signal data point as shown in Figure 3(a). A high DTW value represents inadequacy in model learned features in the structural domain.

**MSE:** Maweu et. al [9] showed that time-frequency domain features can be quantified as an interpretable metric by computing the Mean Square Error between the Continuous Wavelet Transforms (CWT) of any 1D CNN layer and the corresponding 1D data point. CWT (Figure 3(b)) is a well-established feature used in ECG and signal classifiers as an engineered feature. A high MSE value represents poorly learned time-frequency domain features by the model while a low MSE represents a good model.

**SLACK:** Having represented both structural and frequency domain features of ECG 1D signals, clinical features are quantified using the SLACK interpretable explanation metric. ECG signal has characteristic features such as P, Q, R, and S peaks whose inter-distances , heights, frequency, structural, and count of peaks serve as clinical identifiers [9] for heart conditions. While frequency and structural features have already been quantified in the above discussion, the occurrence and count of peaks are quantified using SLACK.

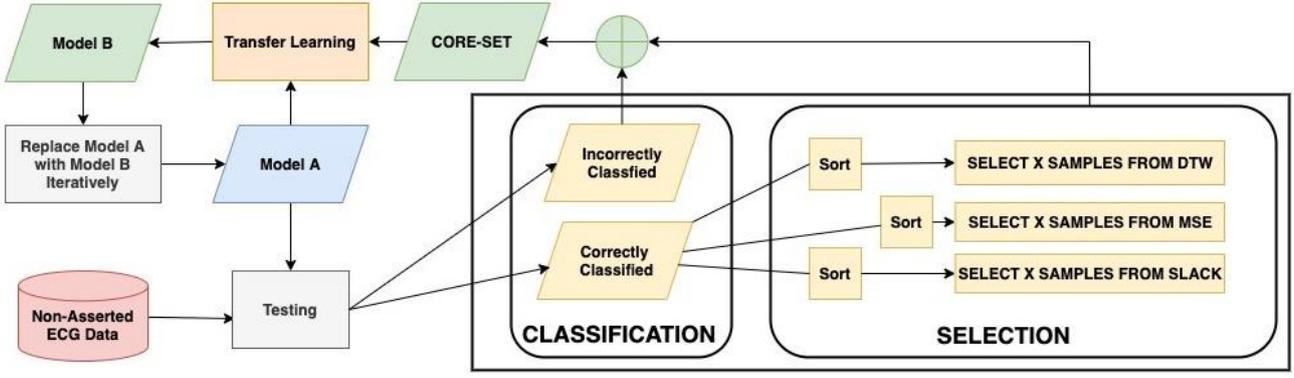

Fig. 4. Explainability-based Core-Set Selection (CSUME) Framework

Maweu et. al focus primarily on the QRS complexes represented by the R peaks as the indicator. SLACK is formulated as shown in Equation 1.

$$SLACK = \left[\left(\frac{\sum_i |RR_{i\_ori} - RR_{i\_pred}|}{RR_{i\_ori}}\right) + \left(\frac{\#RR_{diff}}{\#RR_{ori}}\right)\right] \times 100 \quad (1)$$

where $\#RR_{ori}$ is the count of the number R-R interval observed in the real ECG input signal which includes the count of R peaks. The difference in the count of R-R intervals observed between the real ECG signal and the learned model feature map is quantified as $\#RR_{diff}$. Similarly, the R-R interval value is represented in Equation 1 as $RR_{i\_ori}$ and $RR_{i\_pred}$ represent real ECG signal and feature map at the $i^{th}$ position.

### III. CSUME FRAMEWORK

CEFEs [9] established that explanation metrics DTW, Slack and MSE can provide a threshold and distinguish between classes in a model of adequate capacity. These three metrics allow us to select the samples from the incoming dataset that maximize testing model performance in terms of accuracy, precision, and recall. We also demonstrate data quality assertion for new, incoming, non-asserted ECG data using explanation-based Core-Set selection. Our proposed CSUME framework assumes an existing pre-trained model which is improved upon by transfer learning on our selected samples from incoming data streams iteratively. The new incoming data streams are treated as non-asserted data and are distinct from datasets used in previous iterations. Collected non-asserted data is passed through our *Classification* and *Selection Modules*. An existing *Model $M_A$* is re-trained on the output set of our selection algorithm by transfer learning to obtain *Model $M_B$* as illustrated in Figure 4. We show in Section 5 that the performance on *Model $M_B$* is better than *Model $M_A$* as well as better than a Control Model trained on the entire incoming data without Core-Set selection.

Our proposed methodology identifies samples that maximize the model performance and capacity. Our algorithm has two modules 1) Classification and 2) Selection. Our classification modules use the non-asserted dataset as our test set on existing *Model $M_A$* and group the data as 1) Correctly Classified and 2) Incorrectly Classified. The two groups are treated as independent streams of data.

- Incorrectly Classified: The incorrectly classified data are added to the Core-Set without further filtering. This is motivated by the need to learn better features for the correct classification of these samples which is counter-intuitive to the bootstrapping approach. We assume for our purpose of core set selection that the incorrectly classified group is correctly annotated. We plan to further investigate this assumption in future with a larger ECG dataset.

- Correctly Classified: The correctly classified sample group is passed through our Selection module for filtering.

Our Selection algorithm for the correctly classified is presented in Algorithm 1. The correctly classified samples are subjected to metrics DTW, MSE, and Slack as illustrated earlier. The DTW, MSE, and Slack are computed between the samples and the features maps of the model to be improved. Our algorithm takes the number of classes *(l)* for the task and a user set *budget (b)* as user input. The number of classes is the number of labels for the concerned task such as 2 for a cat or dog classification. While *(l)* is a deterministic constant, the *budget* is a hyperparameter which is a percentage of the new incoming non-asserted data and *(m)* is the number of incorrectly classified samples in this class. We leave the choice of budget to the DL expert's discrimination as it is directly proportional to their need and available computational resource.

$$x = \lfloor (b - m)/3(l) \rceil \quad (2)$$

In Section 4D, we show results with different budget values for illustration. Equation 2 provides a general formula for the calculation of the number of samples to be selected with equal sample distribution to mitigate any class bias. We identify and handle the following edge cases: 1) Non-integer *x* values by rounding to the nearest integer and 2) If the number of samples in a class is less than *x* as calculated by Equation 2, all available samples in the class is selected.

This process is used for the selection of all classes except the class with the lowest number of samples. For the class with the least number of samples, the entire set of samples is selected to mitigate class imbalance when selecting data. Furthermore, the samples are selected without repetition from the set of metrics. This allows our proposed framework to be not only explainable but also bias and class imbalance-free. The selected samples are added to the Core-Set along with

**Algorithm 1: Core-Set Selection**

**Result:** Selected core-set
**Input**: $b, l, m, f_m$, Model Non-asserted ECG data pool
Test Non-asserted ECG data on *Model A*
$CC \leftarrow$ Correctly Classified
$ICC \leftarrow$ Incorrectly Classified
$x \leftarrow \lfloor (b-m)/3(l) \rfloor$
For each Class $C_i$:
  $X_{DTW} \leftarrow$ Calculate DTW between $CC$ and $f_m$
  $X_{MSE} \leftarrow$ Calculate MSE between $CC$ and $f_m$
  $X_{Slack} \leftarrow$ Calculate Slack between $CC$ and $f_m$
  Sort $X_{DTW}, X_{MSE}$ and $X_{Slack}$ in ascending order.
  For each Metric $X_i$:
    $CS \leftarrow$ Select $x$ samples from $X_i$ without replacement
$CS \leftarrow$ Select all samples from $C_l$
$CS \leftarrow$ Append all samples from $ICC$

the incorrectly classified and with the selected correctly classified. When data samples are missing completely or partially for some classes in each dataset, it is still possible to use the CSUME strategy to transfer learn on the DL classifier. This would have to be done based on the judgment of human expert-in-the-loop.

*Evaluation Paradigms:*

To demonstrate the efficacy of our CSUME framework in selection of quality data, we thoroughly evaluate our proposed framework CSUME on six paradigms. In this Section, we define the paradigms used for evaluation, followed by an illustration of the training and testing setup.

- **Paradigm P1 Data Quality Assertion:** We evaluate the CSUME framework's capacity to assert ECG data quality and annotation in Section 4B-1 and show successful assertion of incoming data stream samples.
- **Paradigm P2 Model Improvement:** Having asserted the data quality, we show model performance maximization by the CSUME framework.
- **Paradigm P3 Iterative Model Improvement:** CSUME iteratively maximizes model performance on batches of incoming stream data. In Section 4B-3, we show the results for the iterative model development highlighting that our proposed framework can successfully identify batches of stream data that have a derogatory effect on model performance, thus minimizing the drop in model performance.
- **Paradigm P4 Budget Tuning:** Following the performance results on model improvement, we study the user-defined hyperparameter of our proposed algorithm with different budget values in *20%* increments. Please note the selection of *20%* has no specific reason or significance other than the ease of the representation of results.
- **Paradigm P5 Feature Importance Ablation Study:** We perform an ablation study to identify the order of importance of the selected quantifiable interpretable metrics and their contribution to model performance maximization individually.
- **Paradigm P6 Generalization**: We discuss a generalization of the CSUME framework to model architecture invariance.

### IV. CSUME FRAMEWORK EXPERIMENTS AND RESULTS

#### A. Experiment Setup

*1) Neural Network Architecture A:*

In this Section, we present our experimental results to show performance improvement in the model for our use-case, Arrhythmia classification using the CSUME framework. We evaluate CSUME on six paradigms as illustrated in Section 4B, paradigm P6 on its performance in data sample quality assertion, model performance, iterative improvement, study the effect of our hyperparameter budget and the importance of each quantifiable interpretable metric. The conjugation of both spatial and temporal features of ECG signals makes it challenging for a simple neural network to classify arrhythmia. In this experiment, we used a 13-layer 1D-CNN to train our baseline model with the model architecture consisting of sequentially increasing feature maps followed by Max pooling and Global average pooling as shown in Figure 5. Tanh was used as our activation function for consistently superior results. The model was trained for 100 epochs with 10-fold cross-validation, Adam optimizer, and a learning rate of 0.001.

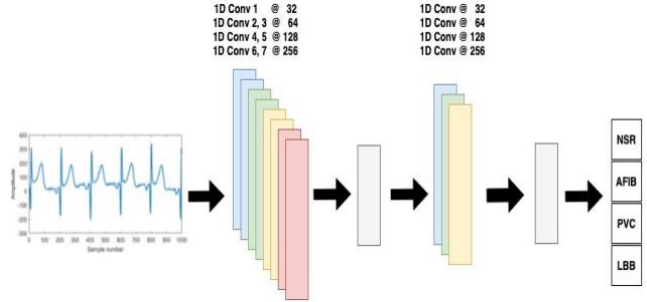

Fig. 5. Convolution Neural Network Architecture for Experiments. The representation "1D Conv 2,3 @ 64" represents that 1D convolution layer 2 and layer 3 have a filter size of 64. The last layer is a softmax over the four output classes NSR, AFIB, PVC and LBB.

*2) Datasets and Experiment Setup:*

We use real ECG recordings provided by [12] derived from 193 MIT-BIH Arrhythmia by [11] as the Asserted Data. The dataset contains 1000 ECG raw 10-beat long signals collected from 45 patients from age 23-89 years for 17 classes of arrhythmia. To show model improvement, we processed the signals to 5-beat long signals and selected 4 of the 17 classes (652 samples) of arrhythmia classes Normal Sinus Rhythm (NSR), Periventricular Contraction (PVC), Atrial Fibrillation (AFIB), and Left Bundle Branch Block (LBB). We split this subset dataset in a 70-30 ratio with 460 training samples as Dataset O, and 192 testing samples as Default Test Set. The Default Test Set is used to evaluate model performance on our experiments in Section 4. The lack of a large-scale public ECG dataset is a challenge in deep learning-based healthcare research. We use generated synthetic data to augment the dataset for proper evaluation. As Non-Asserted data, we used two ECG datasets $S_1$ and $S_2$. Dataset $S_1$ and $S_2$ each have 460 distinct training ECG samples generated using the framework proposed by Maweu et. al [10].

TABLE I. THE ACCURACY, PRECISION, AND RECALL VALUES FOR OUR 1) BASELINE MODELS TRAINED ON ASSERTED DATASET O 2) COMPARATIVE MODEL TRAINED ON A 100% DATASET $S_1$ 3) CSUME BASED SELECTED 98% OF DATASET $S_1$ LEADS TO A 7.81% IMPROVEMENT IN MODEL PERFORMANCE.

| MODEL | TESTING ACCURACY | RECALL | PRECISION |
|---|---|---|---|
| Model $M_A$ trained on Dataset O | 82.81 | 91.75 | 89.47 |
| Model trained on 100% Dataset $S_1$ | 75.52 | 81.00 | 85.51 |
| Model trained on 98% Dataset $S_1$ | 83.33 | 87.39 | 85.89 |

TABLE 2: WE PRESENT THE RESULTS FOR OUR CSUME FRAMEWORK SHOWING MODEL PERFORMANCE IMPROVEMENT USING SELECTION OVER USING THE ENTIRE NON-ASSERTED DATA. WE ACHIEVE A 7.81% ACCURACY IMPROVEMENT WITH A 50% OF INCOMING NON-ASSERTED DATA FOR PARADIGM P2, MODEL $M_B$ AND A DROP IN PERFORMANCE FOR PARADIGM P2, MODEL $M_C$ AND CONTROL MODEL $M_{BC}$ AS ILLUSTRATED IN PARADIGM P3.

| MODEL | ACCURACY | RECALL | PRECISION | BUDGET |
|---|---|---|---|---|
| Model $M_A$ (Dataset O) | 82.81 | 80.95 | 81.93 | - |
| Control Model $M_{AB}$ (Dataset O + Dataset $S_1$) | 86.98 | 82.80 | 92.77 | 100% |
| Model $M_B$ (Core-Set transfer learned on Model $M_A$ with Dataset $S_1$) | 90.62 | 90.62 | 90.62 | 50% |
| Control Model $M_{BC}$ (Dataset O + Dataset $S_2$) | 89.58 | 89.58 | 89.58 | 100% |
| Model $M_C$ (Core-Set Transfer learned on Model $M_B$ with Dataset $S_2$) | 87.50 | 88.42 | 87.50 | 90% |

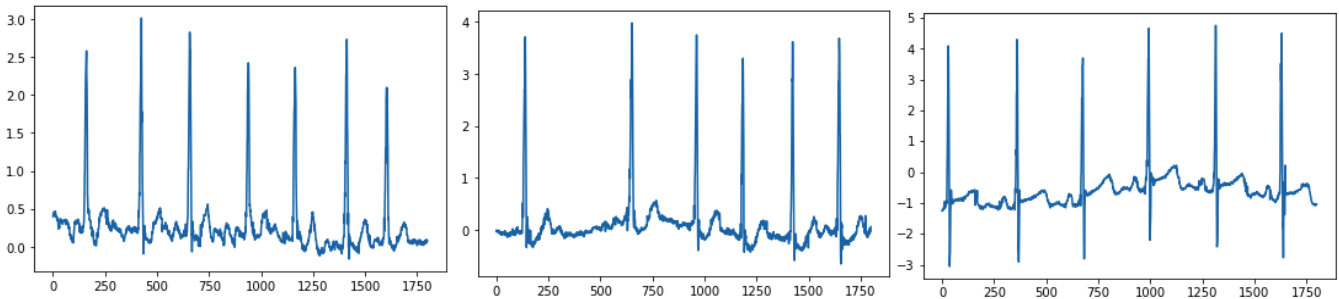

Fig. 6. Sample data points not selected in the Core-Set. X-axis represents time in ms and Y-axis is amplitude.

### B. Paradigm Evaluation Results

#### 1) Paradigm P1: Data Quality Assertion

This experiment shows that our proposed methodology can filter out unwanted samples with high accuracy, therefore asserting the quality and annotation of the remaining samples. To further illustrate this assertion capacity, we trained a model (not transfer learning) with only *98%* (asserted data) of *Dataset $S_1$*, to demonstrate comparable performance to the existing baseline *Model $M_A$* trained on *Dataset O*. Table 1 shows that with a *98%* Core-Set data *budget* on *Dataset $S_1$*, the model achieved similar performance to the baseline *Model $M_A$* and was superior to the model trained with *100%* (non-asserted) *Dataset $S_1$*. Figure 6 presents some samples that were not selected (rejected) by our Core-Set selection algorithm. The visuals clearly demonstrate the noisy nature of these samples. However, it might not be possible to identify if a given noisy sample is clinically relevant for a non-medical person. For example, whether Figure 6(b) represents noise, extended RR interval or a missing beat can only be analyzed and annotated by a medical professional. Annotation or selection of data by multiple experts to attain consensus is costly and time consuming which motivates the need for an automated framework for filtering data samples that are outliers or have a detrimental effect on the decision support system's performance.

#### 2) Paradigm P2: Model Improvement

As illustrated in earlier Sections our framework assumes as existing *Model $M_A$* to be improved with new incoming non-asserted data. This experiment focuses on model performance improvement from *Model $M_A$* to *Model $M_B$* using our modules as presented in Table 2. We also present results for a control experiment to show that *Model $M_B$* performs better with the selected samples over the model trained on entire incoming ECG dataset *Model $M_{AB}$*. Having set a budget of *50%*, the number of samples to be selected per class is calculated by equation 2 from *Dataset $S_1$*. The selected ECG samples from each class per metric, along with the entire of incorrectly classified ECG samples from our Core-Set. Table 1 presents the testing results on our *Default Test Set* for the existing baseline *Model $M_A$*, *Dataset O*, the *Model $M_B$* trained with Core-Set selected from *Dataset $S_1$*, and the Control Model transfer learned on with the entire non-asserted data without selection. The results show a *7.81%* improvement in accuracy over *Model $M_A$* and *3.64%* improvement over the control model with a *50%* reduction in training set size using our proposed algorithm. We also observed a *9.67%* and *8.69%* improvement in Precision and Recall. Please note the budget percentage in Table 2 represents the percentage of non-asserted incoming sample size used in training model by transfer learning.

TABLE 3: THE ACCURACY, PRECISION, RECALL VALUES FOR OUR ABLATION STUDY REMOVING ONE COMPONENT AT A TIME.

| METRICS USED TO TRAIN MODEL | BUDGET | TESTING ACCURACY | RECALL | PRECISION |
|---|---|---|---|---|
| MSE and Slack | 50% | 89.58 | 89.53 | 89.06 |
| Slack and DTW | 50% | 89.06 | 88.95 | 88.02 |
| DTW and MSE | 50% | 90.10 | 90.05 | 90.0 |

*3) Paradigm P3: Iterative Model Improvement*

While Model Improvement focuses on model performance improvement from *Model $M_A$* to *Model $M_B$*. Table 2 focuses on Model Improvement Iteratively from previously established *Model $M_B$* to *Model $M_C$*. In this case, *Model $M_B$*. is used as our baseline model and *Model $M_C$*. is derived using CSUME methodology. The same neural network architecture is used to show continuous iterative improvement. As illustrated in Section 4-2 non-asserted *Dataset $S_2$* was used in this experiment maintaining the original class ratio. As presented in Table 2, we select a budget of *90%* to show Model Improvement Iteratively. The Core-Set was selected using CSUME methodology from an incoming stream of ECG data with *Model $M_B$* (in the previous Model Improvement) as the baseline for transfer learning. With our selected budget, a drop in accuracy was observed. To further investigate, we trained a control model with *100%* of the incoming non-asserted ECG data and observed that by using the entire data, the performance of this deep learning model deteriorated. This shows that these non-asserted ECG data samples do not have additional contributions to improve the model and are therefore rejected.

Through the evaluation on the *Default Test Set,* this experiment demonstrated the importance of our selection module not only to maximize deep learning model performance with a minimal number of ECG training samples but also in preventing derogatory effects on model performance.

*4) Paradigm P4: Budget Tuning*

We performed an ablation study on our model Improvement task for illustration of the importance of modules in our algorithm. In Table 4 we also present results for the ablation study of our hyperparameter *budget* to allow algorithm developers to optimally select the budget as per requirement. We present the hyperparameter study with *budget* value in sequentially increasing order. The results, as presented, show that with an increase in budget size, the model performance generally improves and achieves stability between *50%* (Table 1) to *98%* (Table 4) with no-change in performance.

The model performance in deep learning is significantly impacted by the quality and annotation of data samples. Issues such as skewness, outliers, and noisy data together can be characterized as data quality. Removal of such samples improves the model performance and in turn asserts the data quality. With the selection module (Figure 4), we identify the most informative ECG data samples to be used in training deep learning models at a fraction of resources required otherwise.

TABLE 4: HYPERPARAMETER BUDGET STUDY

| BUDGET | ACCURACY | PRECISION | RECALL |
|---|---|---|---|
| 20% | 81.23 | 81.23 | 81.23 |
| 40% | 81.25 | 81.25 | 81.25 |
| 60% | 90.62 | 90.62 | 90.62 |
| 80% | 90.62 | 90.62 | 90.62 |
| 98% | 90.62 | 90.62 | 90.62 |
| 100% | 86.98 | 82.80 | 92.77 |

*5) Paradigm P5: Feature Importance Ablation Study*

In this Section, we present an ablation study of our interpretable explanation metrics namely DTW, MSE and Slack. In Table 3, we present our results by removing each metric component individually for training. This allows us to understand the individual importance and contribution of each metric to our methodology. Having established model performance maximization with a budget of *50%* in Paradigm P1, we set our budget to *50%* for this experiment. The number of reduced samples by the removal of one metric is compensated by equally distributing the same number of samples in the other metrics to maintain homogeneity of our ablation study. We observed the following from our ablation study:

Though DTW is slightly more important than MSE and Slack (equally important), this importance is insignificant which justifies equally weighted metrics. This observation falls directly in line with the existing literature knowledge that CNN based deep learning models learn primarily the structural features of multimedia data such as edges in an image. Furthermore, it shows that for 1D signal data, frequency and clinical domain features also have equal importance in class distinction. The ablation study also shows that all three metrics are essential for maximizing model improvement.

*6) Paradigm P6: Generalization*

Deep learning system generalization is usually demonstrated with respect to the independence of task, dataset, and neural network architecture. In this Section, we discuss the generalization capacity of CSUME framework. CSUME uses raw input time-series signal and the learned feature maps of a model to compute the interpretable explanations. The learned feature maps can be extracted from any layer of any 1D CNN model making the framework independent of any specific architecture. For a thorough evaluation of DL architecture invariance, a wide spectrum of models needs to be developed and studied in-depth. To show DL architecture independence, we present a limited study with one 11-layer 1D CNN with 6 convolutional 1D layers, Max Pooling, three 1D convolutional layers, and finally a Global Average

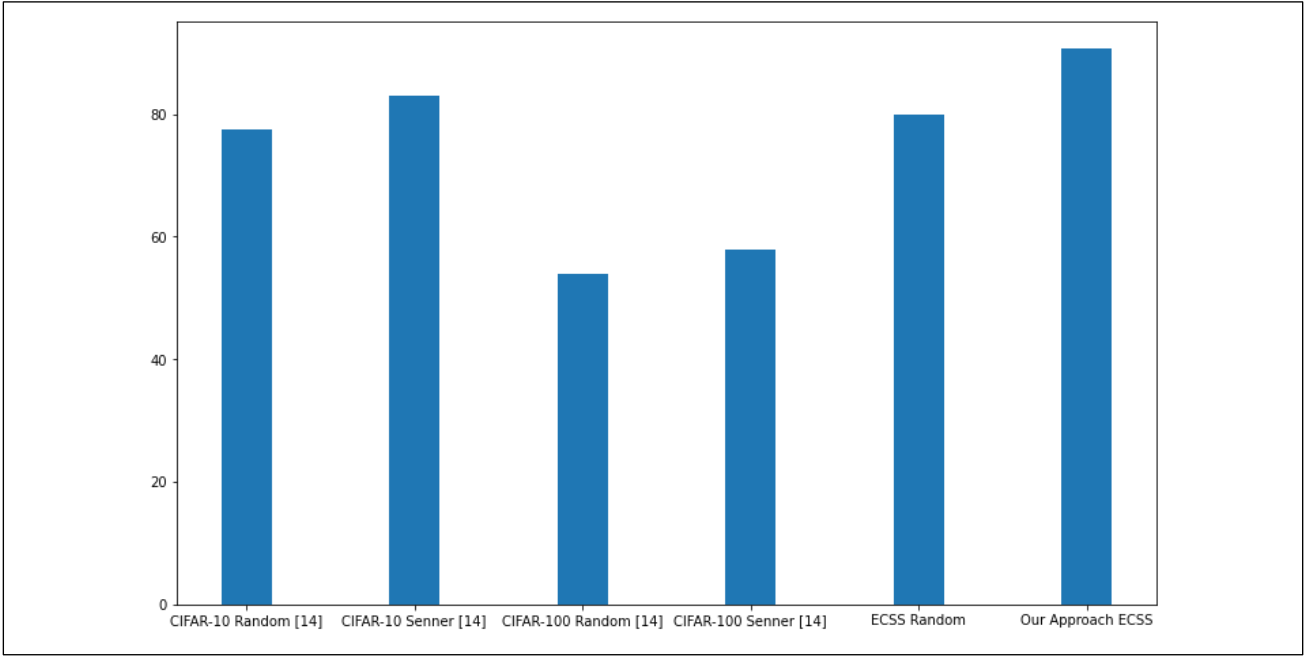

Fig. 7. Core-Set Selection Strategy Comparison

Pooling and refer to it as *Architecture B*. We achieved *81.6%* accuracy with the asserted dataset *(Dataset O)*. Then, we trained on *100%* of *Dataset $S_1$ without* using the proposed CSUME strategy and this resulted in *89.06%* accuracy. Applying the CSUME strategy, the best *budget* value was observed to be *98%* with an accuracy of *92.7%*. This shows that using the *2%* of *Dataset $S_1$* rejected by the CSUME strategy leads to lowering the overall accuracy in the model trained with the entire *(100%) Dataset $S_1$*, thereby demonstrating the ability of CSUME strategy in evaluating individual data sample quality. The *2%* of the rejected samples in this experiment is the same set as the ones reported in Section 4. This also shows that the proposed CSUME strategy can be used effectively in asserting the individual sample data quality before using the data for transfer learning. Ultimately, the degree of accuracy improvement is dependent on the quality of the incoming data, the capacity of the base model used for transfer learning and the budget.

We observe that the degree of model improvement on transfer learning with *Dataset $S_1$* on architecture A is more than on architecture B. We hypothesize that the additional two layers in the architecture A allow for this superior performance. This hypothesis is further bolstered by the observation that architecture A has a *1.21%* better accuracy on the base model trained on *Dataset O* over architecture B. We will need to carry out detailed investigations regarding the effect of additional layers in 1D CNN architectures on the strategies used for Core-Set selection and transfer learning.

## V. COMPARISON WITH EXISTING WORK

In this Section, we compare the results of our proposed CSUME strategy and random selection strategy for our use case with the strategy proposed by [14] on 2D image data CIFAR-10 and CIAFR-100, their corresponding random selection strategy with a data ratio of .5. The lack of 1D healthcare signal Core-Set selection algorithms limit our comparison to existing state-of-the-art research. As observed in Figure 7, with a data ratio of .5, both our proposed strategy and that of Sener et. al improve performance significantly. Although, we observe that the percentage improvement of our strategy is more than that proposed in [14]. It is not a direct comparison due to the different data formats and dataset sizes. Our strategy also allows for an interpretable explanation to the selection of data samples from a dataset for deep learning performance improvement and consequently model degradation.

## VI. CONCLUSION

Deep learning (DL) algorithms require large amounts of high-quality data for providing accurate decision-support. Our proposed CSUME framework uses metrics-based explanations to select the most informative ECG samples to improve DL model performance. The interpretable explanations help understand the effect of individual samples on model performance. Leveraging these quantified explanations, we select the samples to maximize model performance and consequently remove samples that have a derogatory effect on model performance. These metrics-based explanations are used only by the algorithm developer for model improvement and are not intended for end-users. As presented in Section IV, we thoroughly evaluated our proposed framework on 6 paradigms asserting the quality of individual samples, rejecting the potentially noisy or wrongly annotated data, thus consequently improving performance iteratively. Our model improvement results show a 7.81%, 9.67%, and 8.69% improvement in model accuracy, precision, and recall respectively with a 50% reduction in training data. In Paradigms P4 and P5, we analyzed our hyperparameter

*budget* and the importance of each interpretable explanation feature. We discussed the generalization capacity of the framework in terms of deep learning architecture generalization in Paradigm P6 and finally compare our results to Sener et. al [14] showing our strategy improves model performance more than the existing strategies and random strategies. Our strategy also has the additional benefit of providing an explanation as to why a data sample was selected from a dataset for improving deep learning performance and consequently model degradation. The CSUME methodology leverages metrics-based explanations and probably can be used with other appropriate quantifiable interpretable explanations. As future work, we plan to explore metric-explanation based core-set selection for other data modalities and study the efficacy of explainability based core-set selection systems with human-in-the-loop expert intervention. We also plan to investigate the effect of wrong annotations on the model performance and core-set selection. Furthermore, we plan to study the generalization of CSUME framework across various deep learning architectures.